\documentclass{OAGM}
\OAGMarXiv{1505.01065}

\usepackage{graphicx}

\title{Tunnel Surface 3D Reconstruction from Unoriented Image Sequences}

\author{Arnold Bauer$^1$ \and Karlheinz Gutjahr$^1$ \and Gerhard Paar$^1$ \and Heiner Kontrus$^2$ \and Robert Glatzl$^2$ \\ \\
 $^1$JOANNEUM RESEARCH, Graz, Austria \\
 $^2$Dibit Messtechnik GmbH, Innsbruck, Austria}

\begin{document}
\maketitle

\begin{abstract}
The 3D documentation of the tunnel surface during construction requires fast and robust measurement systems. In the solution proposed in this paper, during tunnel advance a single camera is taking pictures of the tunnel surface from several positions. The recorded images are automatically processed to gain a 3D tunnel surface model. Image acquisition is realized by the tunneling/advance/driving personnel close to the tunnel face (= the front end of the advance). Based on the following fully automatic analysis/evaluation, a decision on the quality of the outbreak can be made within a few minutes.
This paper describes the  image recording system and conditions as well as the stereo-photogrammetry based workflow for the continuously merged dense 3D reconstruction of the entire advance region. Geo-reference is realized by means of signalized targets that are automatically detected in the images.
We report on the results of recent testing under real construction conditions, and conclude with prospects for further development in terms of on-site performance.
\end{abstract}

\section{Introduction}

Since almost two decades, 3D scanning has been a method to survey tunnel surfaces, providing sufficient data to be able to reconstruct a tunnel's true geometry with a geometrical resolution of up to 1 point/cm$^2$. The main products to be provided to the various parties involved in tunnel construction are a 3D reconstruction of tunnel vault and tunnel face for geometrical assessment (profile checks) and also true color documentation of geological situation/conditions at the time of excavation. 
Various scanner technologies have been developed and improved since, such as the {\em classic Dibit photo scanner~\cite{PUB06DIB007}}, or laser scanners with up to spherical fields-of-view.

During tunnel advance, time is limited and the environment is hazardous. To be able to quickly assess the success of a tunnel blasting round, the industrial environment demands for efficient, robust and accurate 3D measurements of the gained rock surface. 
%FINAL%Within the K-projekt Vision+ researchers from JOANNEUM RESEARCH together with the industrial partner Dibit Messtechnik GmbH, 
%FINAL%an Austrian provider of leading-edge survey and mapping of linear edifices (tunnels, roads, pipelines, channels), 
The authors have laid the foundation for an automated end-to-end (from image to 3D Model) 3D tunnel reconstruction system, that is purely based on a single camera. Acquiring two profiles of images from two different positions gains a set of overlapping stereo couples (Section~\ref{Image-Capture}). The goal is a dense 3D reconstruction by stereo photogrammetry / structure from motion algorithms making use of these images (Section\ref{Main-Workflow}). To align the reconstructed data set, for each measurement 4-6 signalized targets with known 3D coordinates (measured by an automatically searching total station) are placed on the surface at strategically meaningful positions, those are to be detected automatically by the alignment process. 

\section{Image Recording Under Production Conditions}
\label{Image-Capture}
Due to the nature of excavation, the working area is highly hazardous to spontaneous rockfall or spalls and other dangers caused by destruction of the rock's natural stability. To regain this stability, various means of support have to be set in place by tunnelling construction personnel (miners) such as spikes, wire mesh, lattice girders and as last covering measure, shotcrete to fully cover the rocks natural surface.
As these supporting measures are to be built-in as soon as possible after excavation and mock-out, the complete scanning process must be performed within only a couple of minutes, including setup, data acquisition and geodetic measurements for geo-reference. 
The applied shotcrete fully covers the rock surface: Later re-scanning is impossible.
To cope with these limiting conditions, capable scanning systems must fulfill a set of competing requirements such as {\em easy-to-use, light, mobile, easy to maintain, and cheap}.

Various challenges to data acquisition itself and to later data processing must be overcome to get not only usable but best possible results from the complete process. Recording of scan data, either directly in 3D or as images, in a tunnel environment under production conditions is determined by various constraints such as 
the  dependency from resources (light, power supply),
geo-referencing (position and orientation determination) of scan data,
communication for data transfer,
environmental influences (dust, damp, dripping water, dirt),
ruggedness of hardware,
mobility (one person can easily carry the equipment),
personnel (operators),
ease of handling (simple, structured), or
field-of-view (the number of image sequences should be minimized).

\begin{figure}[ht]
  \centering\includegraphics[width=\linewidth]{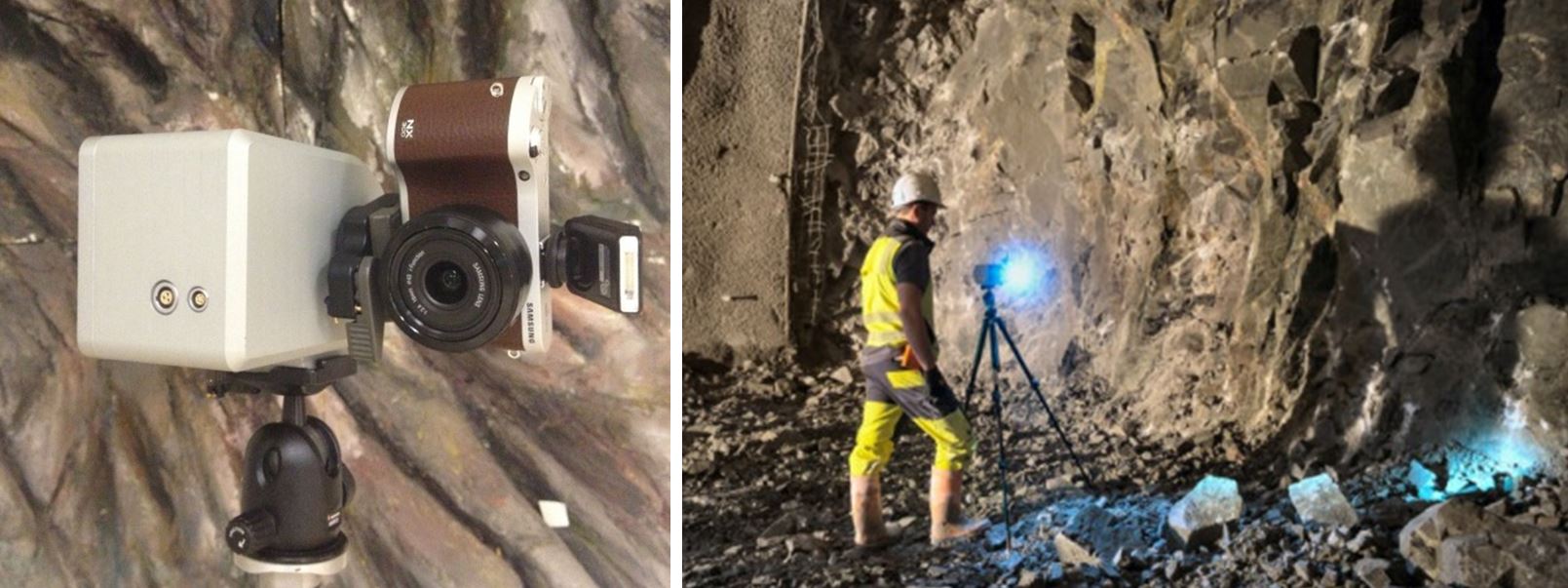}\\
  \vspace{-2mm}
  \caption{Left: 
%FINAL% Dibit Handheld data
Data acquisition system based on a consumer camera with fixed focus lens, and a stepper motor unit. Right: Typical data acquisition situation close to the hazardous region of the tunnel face.}
  \label{fig:ImageCapture}
\end{figure}

\begin{figure}[h]
  \centering\includegraphics[width=\linewidth]{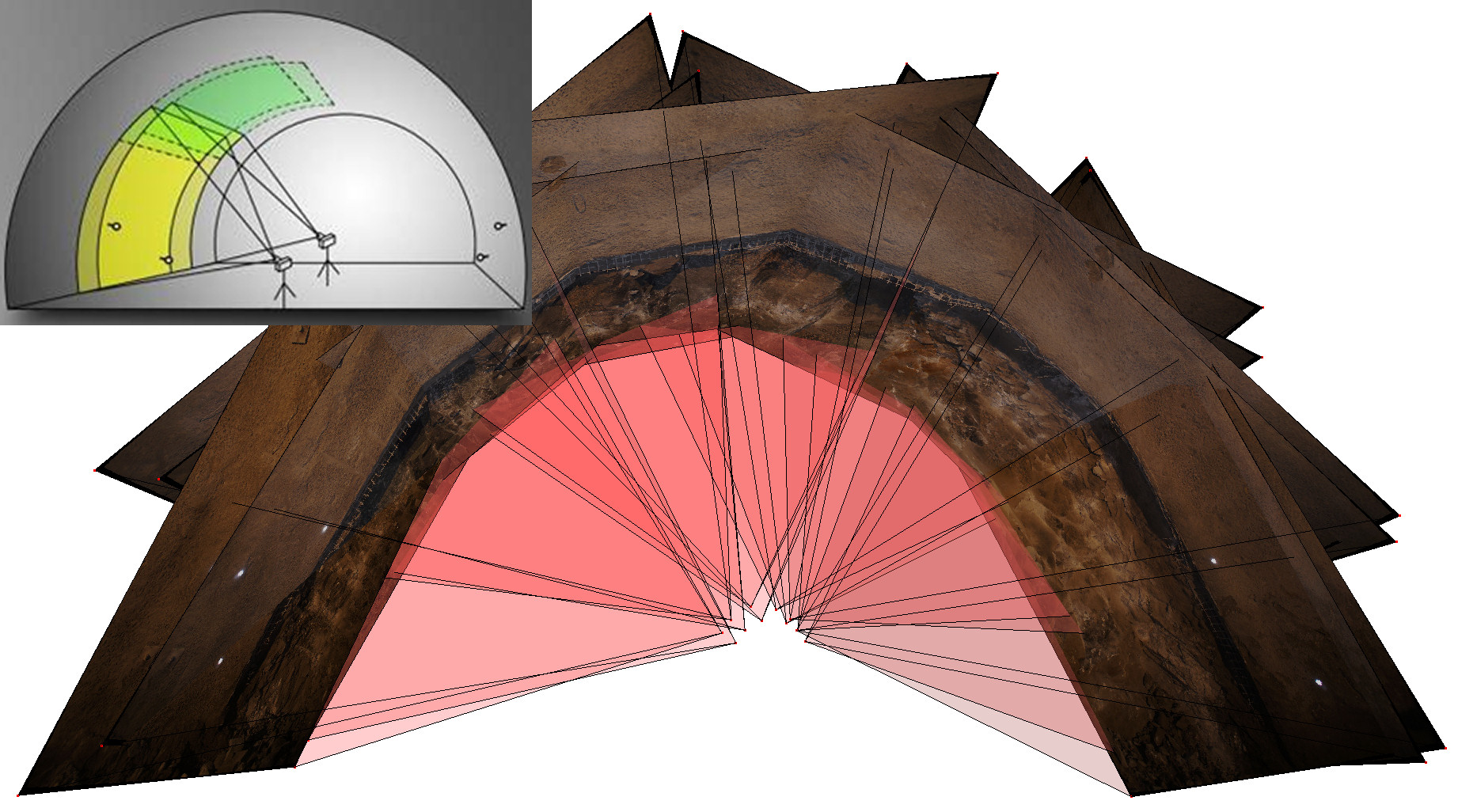}\\
\vspace{-2mm}
  \caption{Image footprints in 3D space (top left: Image acquisition geometry).}
  \label{fig:ImageGeometry}
\vspace{-2mm}
\end{figure}

\begin{figure}[h]
  \centering\includegraphics[width=\linewidth]{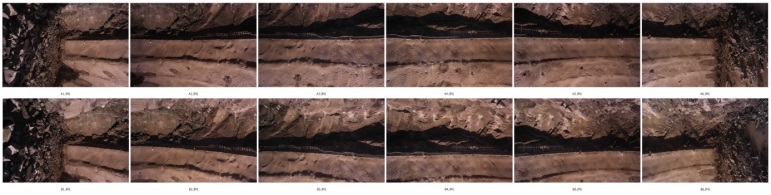}\\
  \vspace{-2mm}
  \caption{Left: Two stereo-overlapping profiles of images of the tunnel surface.}
  \label{fig:StereoOverlapping}
\end{figure}

For simplified handling, development targeted towards a single-camera based imaging device equipped with a motorized tilt mechanism to acquire a 360$^{\circ}$ (vertical) panorama.
In its typical scenario the system is placed on a tripod close to the tunnel face in two slightly distinct positions (Fig.~\ref{fig:ImageGeometry}) by one on-site staff person (Fig.~\ref{fig:ImageCapture}). This way, stereoscopic panoramas are available. Fig.~\ref{fig:StereoOverlapping} depicts an example of such two panoramas, forming stereo partners usable for further 3D reconstruction. Processing complexity constraints require the user to obey some simple rules for the recording process: For positioning and alignment the scanning system is equipped with a laser-pointer aid. In addition, the system should be set up roughly horizontal, with no need for further exact leveling. 
The light source should be positioned approximately one tunnel diameter behind the tunnel face and should not change during image acquisition. Two targets need to be positioned at each side of the tunnel.
Reference measurements of the control point targets are performed remotely by an automatic total station. The total station is controlled by a rugged field computer which is also used for quick data processing and the geometry assessment (profile checks) and excavation optimization by the construction personnel within the tunnel.

\section{From Unoriented Images to 3D Tunnel Model}
\label{Main-Workflow}

From the construction companies' point of view, the excavated areas shall not only be documented, but also in-situ assessed. The scanning data must be processed typically within  three to five minutes to an easy to understand overview about the tunnel's true geometry for immediate optimization of construction works. As construction personnel is neither acquainted with surveying tasks nor data processing, this complex task requires a {\em fully automatic and reliable workflow for data processing}, covering remote target registration by automated total stations, target recognition and 3D reconstruction during the processing workflow and last, presentation of the results in an easy to understand depiction of the assessed tunnel section. After data acquisition and quick processing in the tunnel, further data processing, merging with existing or preceding scanning data and quantity survey is usually carried out in the field office by surveying or construction engineers. Geological or geotechnical interpretation of the excavated rock is further made by the experts in the respective fields. 

\begin{figure}[h]
  \centering\includegraphics[width=\linewidth]{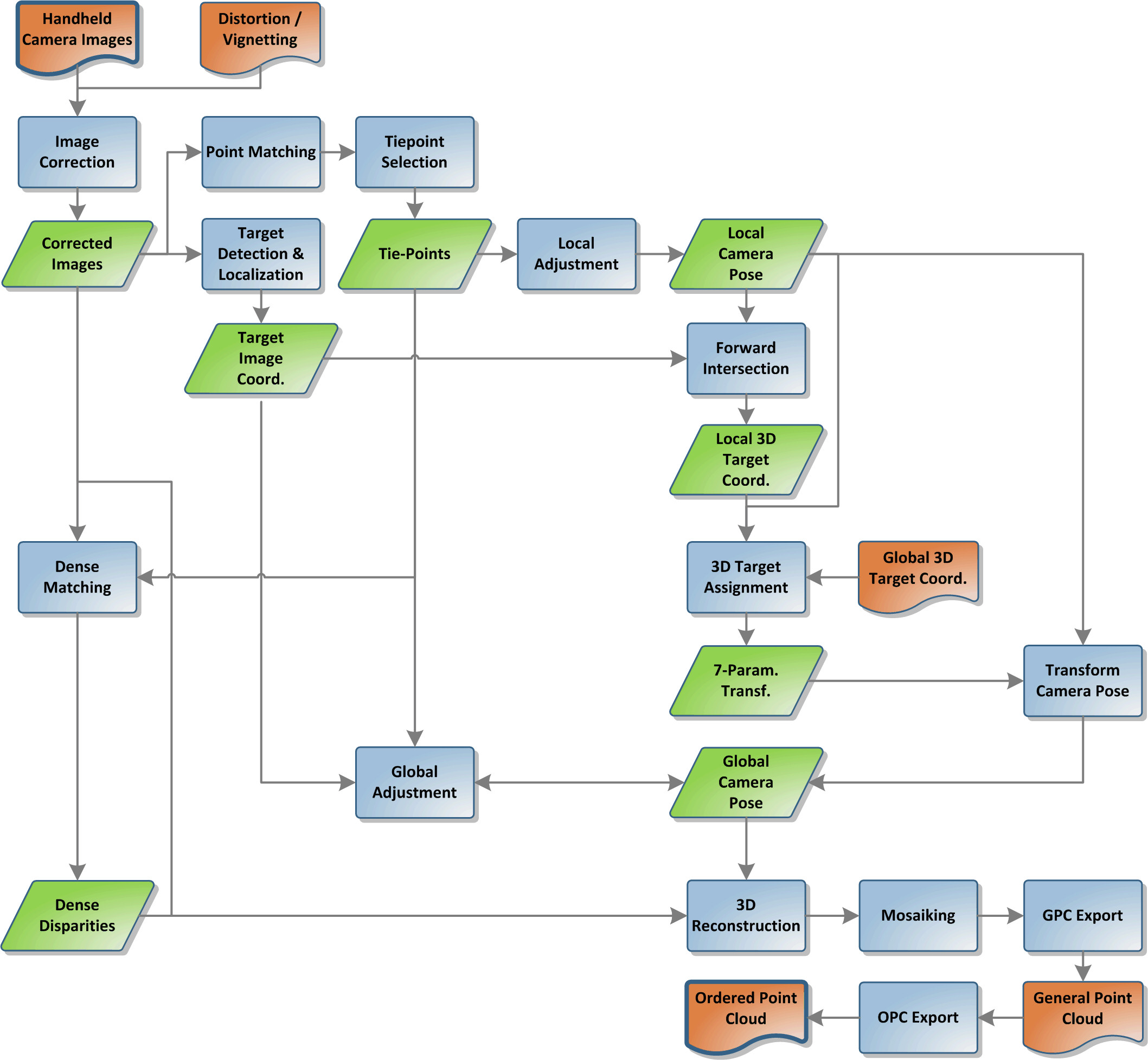}\\
  \caption{Workflow overview }
  \label{fig:Workflow}
\end{figure}

In such a scenario, compared to controlled industrial (production-like) conditions, some challenges are posed to the processing framework, such as unordered image sequences, fairly unpredictable stereo base-length, inhomogenous scene illumination, partial under- or overexposure, re-positioning of light source during image acquisition, or unexpected objects (persons or equipment obstructing the tunnel surface) appearing in the images or casting shadows. 
In the following the main steps  (Fig.~\ref{fig:Workflow}) as combined into a Python-based test suite are described.

%\subsection{Image Preprocessing}
{\bf \underline{Image Preprocessing}} consists of radial and tangential distortion correction and vignetting correction.
%Image geometry correction (distortion  as deviation from rectilinear projection) is compensated by image warping based on radial and tangential distortion coefficients derived %from camera calibration.  Vignetting (a decrease in brightness in image edges when compared to the center) is compensated using camera-specific radiometric calibration %images. 
In order to speed up processing image size is optionally reduced for fast on-site reconstruction. 

\begin{figure}[ht]
  \centering\includegraphics[width=\linewidth]{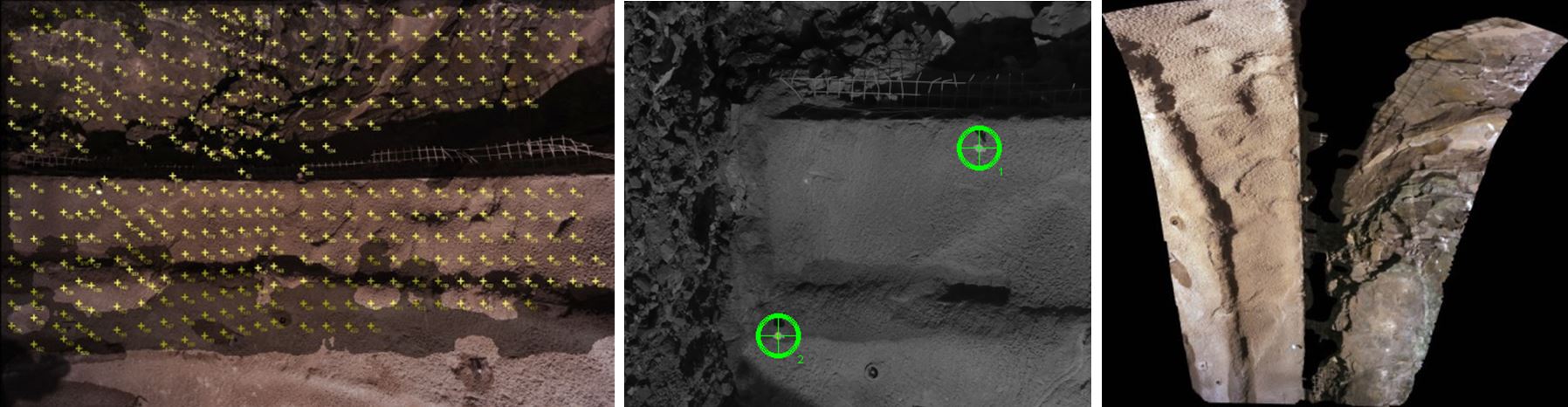}\\ 
\vspace{1mm}
  \centering\includegraphics[width=\linewidth]{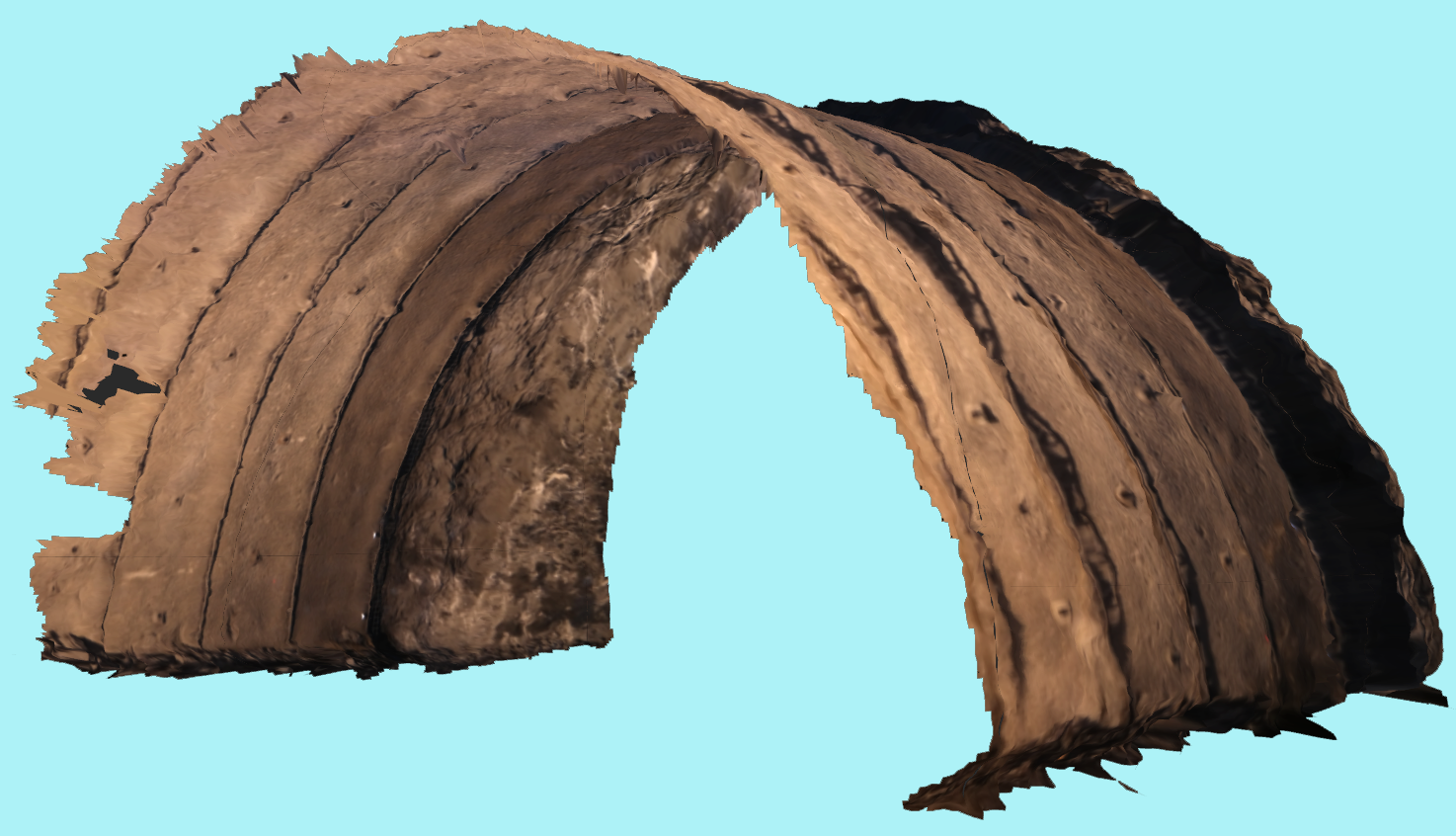}\\
  \caption{Examples for intermediate results of the
%FINAL% Dibit 
Handheld workflow Top left: Tie points used for bundle adjustment. Top middle: Detected and localized signalized points. Top Right: Single-stereo 3D reconstruction projected on a spherical coordinate space. Bottom: Entire profile visualized in 3D.}
  \label{fig:Examples}
\end{figure}

%\subsection{Tie-pointing}
{\bf \underline{Tie-points}}
(corresponding points in adjacent images) 
%are necessary for the following bundle block adjustment to derive mutual image orientation.
%Local feature detection is used to identify distinctive interest points. 
%The neighbourhood of each interest point is described by a feature vector. This descriptor is matched between different images to find correspondences. 
are identified using a derivate of Speeded Up Robust Features (SURF)~\cite{Bay2008} (Fig.~\ref{fig:Examples} left). Adaptive histogram equalization and image tiling allows to maximize coverage of tie-points, to ensure proper distribution, and to be insensitive to different texture classes (e.g. raw rock and shotcrete). For finding correspondences feature vector matching is used which compares Euclidian distance, incorporating best to second-best match ratio, nearest neighbour approximation based on FLANN~\cite{flann_pami_2014}, geometric constraints, and outlier detection.

%\subsection{Local Bundle Block Adjustment}
%\label{Bundle-Block-Adjustment}
{\bf \underline{Local Bundle Block Adjustment}}
uses the complete tie point set to generate a locally consistent set of orientations for all images involved. Since camera pose itself is not directly measured and transformation in reference project coordinate system is done by inspecting targets within the tunnel region-of-interest, modelling is split into two adjustment steps. {\em Local adjustment} results in a camera pose estimation which is consistent but imprecise and has wrong scaling. This already allows calculating local 3D target coordinates. {\em Global} 3D target coordinates are externally provided (by standard theodolite tunnel geodesy infrastructure), therefore target assignment has to be resolved later. 
%Matching local and global coordinate sets (being an implicit step of global adjustment, see below) allows identifying the correct point mapping automatically.

The used local and global bundler is based on the well-known (linearized) co-linearity equations~\cite{Luhmann2010Nahbereichsphotogrammetrie}. The adjustment is described by a generalized Gauss Markov model~\cite{Kariya2004Generalized} but we use the fact that the design matrix holding the derivatives with respect to the measurements is quadratic and non-singular leading the adjustment scheme easily to be converted to an ordinary least squares scheme. 
To increase the number of unknown parameters, we first reduce the normal equations by eliminating the a priori not interesting 3D coordinates of the tie-points. 
This again can be done as the part of design matrix holding the derivatives with respect to the 3D coordinates of the tie-points has block diagonal structure and each sub-block is quadratic and non-singular. 
After solving the reduced adjustment problem this reduction of the normal equations is inverted and also the updated 3D coordinates of the tie-points are derived.

For the {\em local adjustment} the first camera position is set to zero and the second exactly one distance unit apart. The rotation angles are initialized by assuming a regular sampling along the tunnel surface starting from and ending in a horizontal position. The major challenge of low to no stereo base between images of one rotation cycle could be resolved by inserting additional constraints and checks in the bundle adjustment procedure. We therefore excluded tie-points which were localized only in two or more images of the same tripod position and/or which were localized in less than a specified number of images.
In the local adjustment the camera positions were kept fixed yielding only a relative oriented model. The adjustment is an iterative procedure, as the 3D coordinates have to be updated using the current orientation parameters. We additionally use the stochastic model to avoid an incremental rotation of the relative oriented model around the stereo base.

%\subsection{Signalized Targets Detection and Localization}
{\bf \underline{Signalized Targets Detection and Localization}}
is done by local thresholding, Hough transform, and morphology operations. Additional constraints such as target diameter, distance to camera, circularity tolerance, and position help to minimize false positives. Geometric and epipolar constraints allow identifying and assigning the targets to their 3D coordinates even in regular industrial conditions (Fig.~\ref{fig:Examples}, middle).

%\subsection{Global Orientation}
%\label{Global-Orientation}
{\bf \underline{Global Orientation}}
simultaneously for all images is obtained by a seven-parameter transformation between the known 3D signalized points from total station and their local 3D coordinates as from the local bundle adjustment results and 2D image positions. RANSAC is applied to detect the correct combination.
The fixed camera positions from local adjustment are further optimized. Compared to local adjustment the complexity is relaxed as the exterior orientations' estimates are now already very well predicted. For low to no stereo base between images of one rotation cycle we use the same tie-point elimination strategy as in the local adjustment with a different weighting strategy. 

%Moreover we balance the weight of the very few known targets (serving now as ground control points) with respect to the mass of derived tie-points.
%If the achieved accuracy is below a given threshold we rerun the global adjustment but additionally search for erroneous point measurements. 
%This automated point removal is based on the individual back projection errors. For a large number of measurements the residual threshold can be derived from the overall RMS values. 
%If less measurements are available we used the median absolute deviation (MAD) which is less sensitive to gross errors.

%\subsection{Dense Stereo Matching}
{\bf \underline{Dense Stereo Matching}}~\cite{PUB93DIB008} is based on predictions provided from the tie-points matching and exploits RGB color as individual cue. 

%\subsection{3D Reconstruction and Tunnel Surface Modeling}
{\bf \underline{3D Reconstruction and Tunnel Surface Modeling}} uses disparities for Locus reconstruction~\cite{PUB97DIB007} to project the resulting 3D and textural information on a surface grid on the sphere, resulting in individual patches of radial distances and texture (Fig.~\ref{fig:Examples}, top right).
3D patches are merged in the same spherical coordinate system grid by unsharp masking. The result is a consistent 3D reconstruction of the acquired profile with a typical length of 4m (Fig.~\ref{fig:Comparison}).

\section{Experience and Results}

\begin{figure}[ht]
  \centering\includegraphics[width=\linewidth]{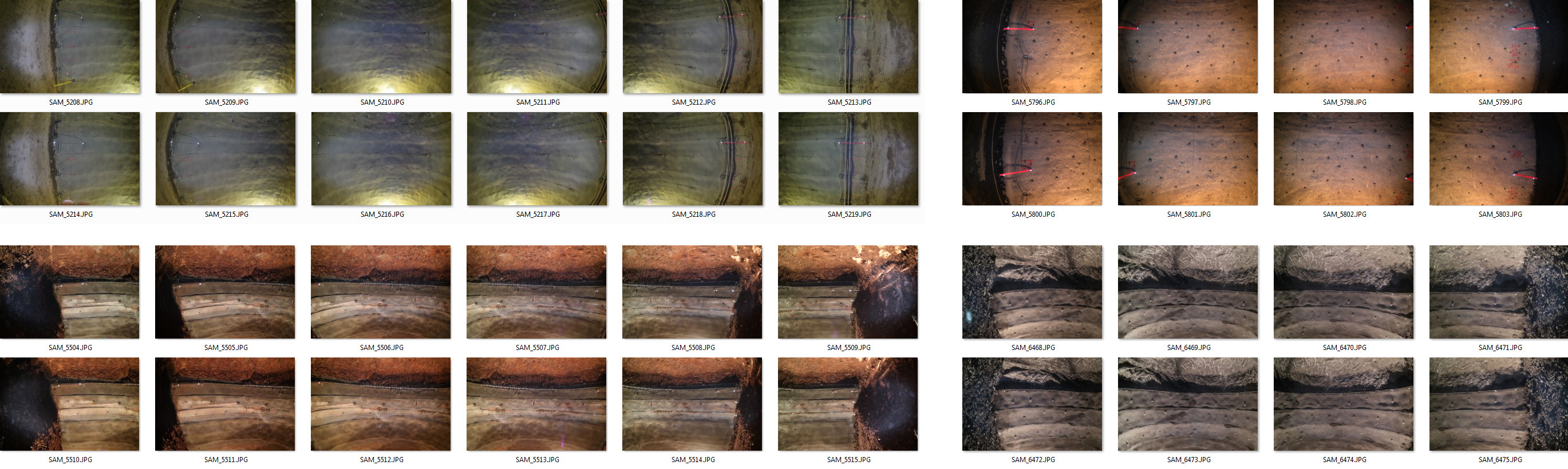}\\
  \centering\includegraphics[width=\linewidth]{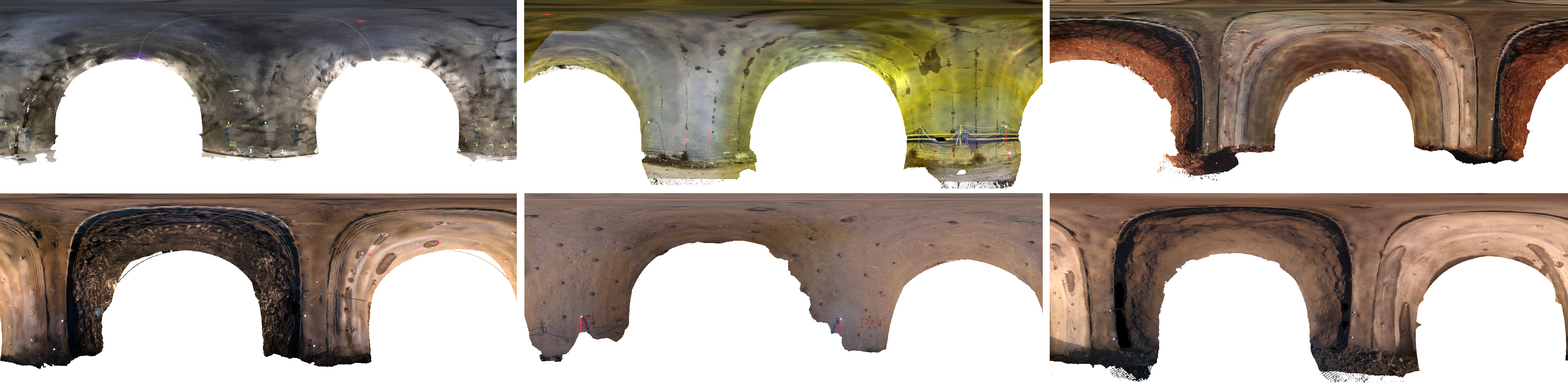}\\
  \caption{Top: Examples of test data sets with different camera setups and positions, different texture and illumination, and delicate environmental conditions. Bottom: Examples of automatically reconstructed 3D surface models (spherical projection) of different data sets.}
  \label{fig:Testdata}
\end{figure}

\begin{figure}[ht]
  \centering\includegraphics[width=\linewidth]{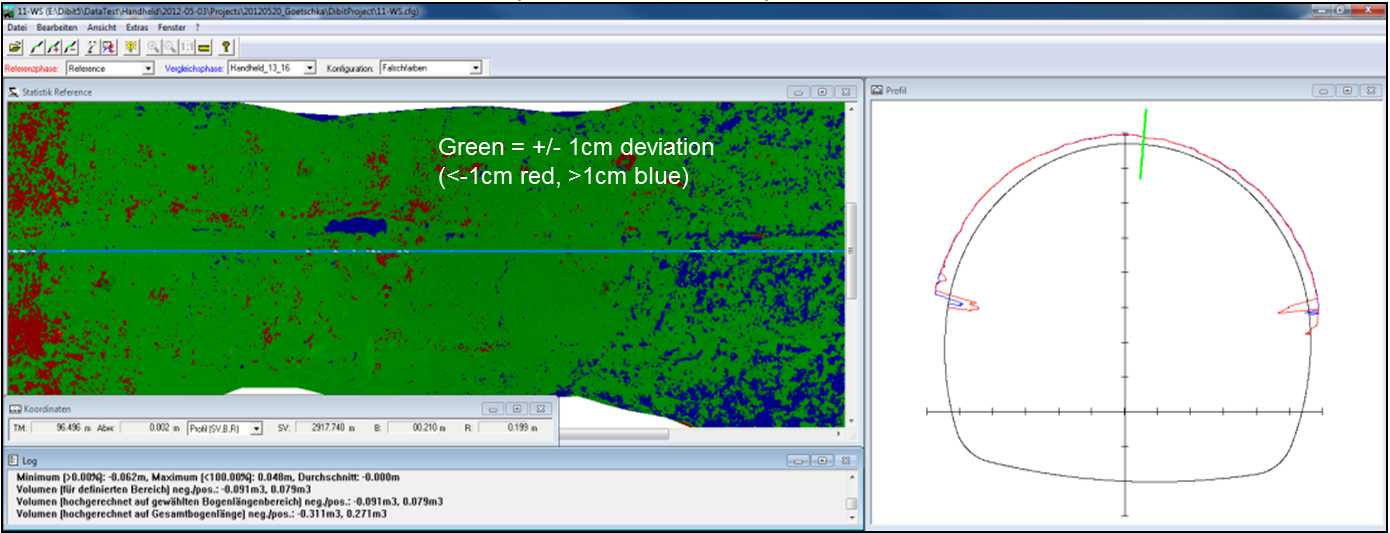}\\
  \caption{Comparison of 
%FINAL%Dibit 
Handheld result with georeferenced laser scan of the same area. Left: Color-coded deviation on unwrapped tunnel model. Right: Profile of laser-based and 
%FINAL%Dibit 
Handheld reconstructions superimposed }
  \label{fig:Comparison}
\end{figure}

To obtain a stable, reliable system, intensive testing was conducted (unit testing, integration testing, interface testing, and acceptance testing). Using a 24mm wide-angle lens with 82$^{\circ}$ field-of-view, 6 images are acquired to cover the whole tunnel profile with adequate image overlaps. Full resolution is 20 MPixels per image. The application is focussed on fast processing providing results promptly after measuring with reduced image size of 1 MPixels and evaluation time of few minutes on standard hardware. Dependent on the overlap area the typical number of tie-points is 30-100 per image. Extensive field-tests showed that it is possible to reduce signalized targets to 4, always 2 on opposite tunnel side. We optimized point detection and matching parameter sets for different textures  (e.g. rock, shotcrete) and construction stages (raw excavation, tunnel face, top heading, bench). 34 representative data sets were provided by 
Dibit Messtechnik
%the industrial application 
(see Fig.~\ref{fig:Testdata}, top for a subset) including also problematic and low-quality data, e.g. including occlusions due to moving objects, changing illumination, shadows, dirty lens, and critical deviation from requirements. At the end of the test cycle the software was able to successfully handle all the data sets (Fig.~\ref{fig:Testdata}, bottom) and to explore the possible limits of the Handheld  system. Comparisons with geo-referenced laser scans of the same region show deviations less than 10mm  (Fig.~\ref{fig:Comparison}) which is well within the requirements during tunnel advance.

\section{Conclusions and Outlook}

The system is currently under extensive testing in real conditions on several construction sites within Europe. It is designed to be operated by regular staff as available during tunnel advance, and will not need expertise in computer vision, photogrammetry, or surveying. Costs are only those for a regular consumer camera with appropriate lens, and the simple motorized tilt mechanism. Once introduced in the regular construction process, it will significantly reduce the costs of tunnel advance documentation and at the same time raise the richness of the gained data: Compared to state-of-art laser scanning hardware cost reductions of several tens are reached with similar performance, accuracy and data quality. Data acquisition time (effecting the advance process) is minimized to a few minutes. Safety for the operating staff is ensured by placing the measurement system out of the hazardous region in the immediate vicinity of the tunnel face.
Less staff presence, lower hardware costs, lower interruption periods during actual construction works, 
higher safety for operating staff and the immediate availability of a full 3D description of the recorded site (in all desired stages) 
make the 
%FINAL%Dibit 
Handheld system a new high-value component of the New Austrian Tunneling method, 
%originated by Austrian funded research and development and fed by novel 3D vision mechanisms designed for practical robust application.

Development is still ongoing. Several re-design steps in many parts of the workflow and Hardware set-up have been realized within the last 2 years to make it applicable for practical applications, operated by non-expert staff.
%
%To give some examples, extensive tests showed necessity of the following improvements, which were realized during 2014:
%\begin{itemize}
%\item Reduce to at least 4 targets for geocoding to minimize the additional effort for image acquisition.
%\item Reduce to maximal 4 images for each camera position to cover the whole profile. This implies to reduce the overlapping area between adjacent images to a minimum.
%\item Handle low-resolution (0.3 MPixel) up to high-resolution (20 MPixel) images for close-to real-time to maximum resolution processing.
%\item Reduce camera eccenter (from 25 to 4cm) to allow a very compact camera system construction.
%\item In stereo matching handle different texture types within the same image, e.g. including both excavation and shotcrete with often different illumination
%\item Implement more robust relative orientation for local pose predictions
%\item Reduce the need of input parameters, such as stereo basis, tunnel geometry, and camera position w.r.t. tunnel axis.
%\end{itemize}
%
Current work includes the integration of fisheye lenses to reduce the number of images to be recorded and to take viewpoints further away from the hazardous area close to the unsecured tunnel advance. A CUDA implementation for SGM (Semi-Global-Matching) will be introduced soon. Reducing the number of targets necessary for georeference, alternative orientation strategies as well as immediate feedback on image acquisition success in terms of data usability and scene coverage are further foreseen improvements.

\section*{Acknowledgments}
\vspace{-3mm}
{\small
This work was supported by the K-Project Vision+ which is funded in the context of COMET - Competence Centers for Excellent Technologies by BMVIT, BMWFJ, 
Styrian Business Promotion Agency (SFG), Vienna Business Agency, Province of Styria -- Government of Styria. 
The programme COMET is conducted by Austrian Research Promotion Agency (FFG).
Further funding sources for this work are JOANNEUM RESEARCH Forschungsges.m.b.H. and Dibit Messtechnik GmbH.
%To be given in final version}

\bibliography{bibsource_gut,FER_MVA_Publikationen_2014-12-01,FLANN,SURF}
\end{document}